\pdfoutput=1
\PassOptionsToPackage{dvipsnames}{xcolor}
\documentclass[11pt]{article}
\usepackage[table]{xcolor}

\usepackage{graphicx}
\graphicspath{{./Images/}}
\usepackage{multirow}
\usepackage{comment}
\usepackage{hyperref}
\usepackage{amsmath}
\usepackage{amssymb}
\usepackage{bbm}
\usepackage{makecell}
\usepackage{listings}
\usepackage{tabularray}
\usepackage{colortbl}
\usepackage{soul}
\usepackage{subcaption}
\usepackage{pifont}

\newcommand*\colourcheck[1]{%
  \expandafter\newcommand\csname #1check\endcsname{\textcolor{#1}{\ding{52}}}%
}

\newcommand{\hlc}[2][yellow]{{%
    \colorlet{foo}{#1}%
    \sethlcolor{foo}\hl{#2}}%
}
\usepackage[final]{acl}

\usepackage{times}
\usepackage{latexsym}

\usepackage[T1]{fontenc}

\usepackage[utf8]{inputenc}

\usepackage{microtype}

\usepackage{inconsolata}

\title{Merging Facts, Crafting Fallacies: Evaluating the Contradictory Nature of Aggregated Factual Claims in Long-Form Generations}

\author{Cheng-Han Chiang \\
  National Taiwan University,\\ Taiwan\\
  \texttt{dcml0714@gmail.com} \\\And
   Hung-yi Lee \\
  National Taiwan University,\\ Taiwan \\
  \texttt{hungyilee@ntu.edu.tw} \\}

\begin{document}
\maketitle
\begin{abstract}

Long-form generations from large language models (LLMs) can contain a mix of factual and non-factual claims, making evaluating factuality difficult.
Prior works evaluate the factuality of a long paragraph by decomposing it into multiple facts, verifying those facts independently, and aggregating the results.
Such methods assume that combining factual claims forms a factual paragraph.
The above assumption can be violated: we show that strong open-source models like Llama-chat can generate paragraphs that contain verifiable facts, but the facts are combined into a non-factual paragraph due to entity ambiguity.
We further reveal that existing factuality metrics, including FActScore and citation recall, cannot properly evaluate these non-factual paragraphs and overestimate their factuality.
To address this, we introduce an enhanced metric, \textbf{D-FActScore}, specifically designed for content with ambiguous entities.
We evaluate the D-FActScores of people biographies generated by retrieval-augmented LLMs.
We show that D-FActScore can better assess the factuality of paragraphs with entity ambiguity than FActScore.
We also find that four widely used open-source LLMs tend to mix information of distinct entities to form non-factual paragraphs, making their D-FActScore much lower than FActScore by over 10\%.

\end{abstract}

\colourcheck{ForestGreen} 
\colourcheck{Bittersweet}
\colourcheck{NavyBlue}

\section{Introduction}

LLMs can generate high-quality texts, making LLMs prevalent in everyday usage~\citep{ChatGPT,openai2023gpt4}.
However, LLM's generation may not always have a high \textit{factual precision}~\citep{nakano2021webgpt,rae2021scaling}.
Factual precision measures whether the information conveyed in the text is factually accurate.\footnote{We only focus on factual precision and do not consider \textit{factual recall}, i.e., how well the generation covers the information. This paper will use \textit{factuality} to refer to factual precision.}
As long-form generations can contain a mix of factual and non-factual claims, recent works propose to evaluate the factuality of long-form generation in a more fine-grained way that considers the factuality of each claim in the generation.
Precisely, these methods decompose the generation into several claims (called \textit{atomic facts} in~\citet{min-etal-2023-factscore}) and then verify each claim independently.
The factuality of the whole generation is the percentage of verifiable claims.
Widely used evaluation metrics like FActScore~\citep{min-etal-2023-factscore} and citation recall~\citep{liu-etal-2023-evaluating,gao-etal-2023-enabling} fall into this type.

\begin{figure}[t]

\centering
\includegraphics[clip, trim = 20px 25px 22px 15px,width=1.0\linewidth]{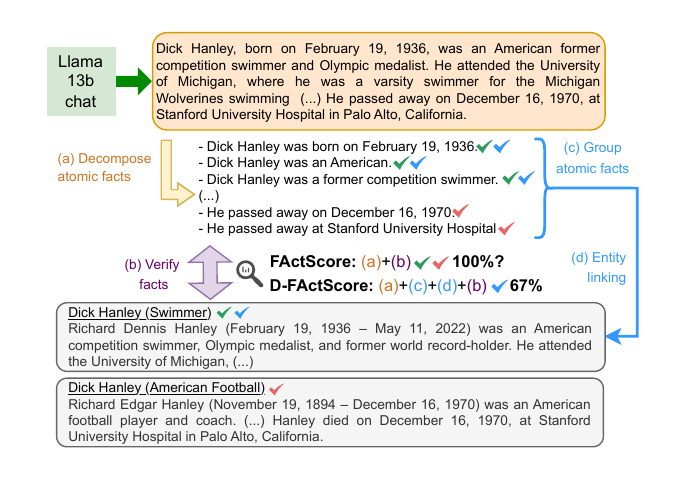}

\caption{Output of Llama-13b-chat when prompted to generate a biography for Dick Hanley.
While the paragraph is misleading and non-factual, all the facts in the paragraph can be supported by the Wikipedia of \texttt{Dick Hanley (Swimmer)} \ForestGreencheck\:or \texttt{Dick Hanley (American} \texttt{FootBall)} \Bittersweetcheck, yielding 100\% FActScore.
D-FActScore groups atomic facts that \textit{appear to} refer to the same individual based on the paragraph (Figure~\ref{fig:illustration.pdf}\textcolor{NavyBlue}{(c)}), finds an entity that best matches that individual from the knowledge source (Figure~\ref{fig:illustration.pdf}\textcolor{NavyBlue}{(d)}), and only uses the information of that entity to verify the facts in that group \NavyBluecheck.
}
\label{fig:illustration.pdf}
\end{figure}

The correctness of the previous factuality metrics relies on the assumption that "\textit{as long as each claim is factual, the combination of those claims is factual}."
This paper shows that this assumption can be violated due to entity ambiguity in the generation.
Figure~\ref{fig:illustration.pdf} shows such a case, where Llama-2-13b-chat says that Dick Hanley was born in 1936 and died in 1970, where the birth date is true for \texttt{Dick Hanley (Swimmer)} and the time of death is true for \texttt{Dick Hanley (American Football)}, but readers will think all the information refers to the same individual.
Since the paragraph is not written in a way that allows readers to understand the content factually, the paragraph is non-factual.
However, since each claim in the generation is supported by Wikipedia, the FActScore of the above misleading paragraph is 100\% by definition.

We emphasize the significance of this evaluation problem by showing that LLMs can easily generate this kind of non-factual paragraph.
We collect 500 ambiguous names from Wikipedia to form a dataset, AmbigBio, where each name corresponds to multiple entities in Wikipedia.
We prompt LLMs to generate a biography for a name in AmbigBio using retrieval-augmented generation~\citep{guu2020retrieval,lewis2020retrieval}, where the target name is shared by multiple entities, and the retrieved documents may also include the Wikipedia pages of multiple entities.
We find that Llama-13b/70b-chat~\citep{touvron2023llama}, Tulu-13b-dpo~\citep{ivison2023camels}, and Vicuna-7b~\citep{vicuna2023} tend to mix the information of multiple entities in the same biography, and a reader without prior knowledge cannot tell that the information corresponds to different entities.
Non-factual as these paragraphs are, existing factuality metrics cannot correctly assess the factuality in this case.

To solve this evaluation problem, we modify FActScore into \textbf{Disambig-FActScore} (D-FActScore), which handles entity ambiguity better.
Unlike FActScore, which verifies each atomic fact in the paragraph independently, D-FActScore splits the facts in a paragraph into groups.
If the \textit{narration} of the paragraph can make a reader without prior knowledge think that the two facts are about the same individual/object, they belong to the same group.
Next, D-FActScore uses entity linking to find an entity in the knowledge source that best matches a group of facts and use this entity to verify the facts in the same group.

We recruit humans to annotate the D-FActScores of people biographies generated with Llama-13b-chat, Tulu-v2-13b-dpo, and ChatGPT~\citep{ChatGPT}.
We show that Llama and Tulu mix the information of multiple entities in a non-factual way. 
Compared to FActScore, D-FActScore correctly identifies those non-factual generations and gives a D-FActScore 8\% to 16\% lower than FActScore. 
We also propose a pipeline to estimate D-FActScore automatically and show that our automatic pipeline estimates D-FActScore to a reasonable degree.

Our contributions are summarized as follows:
\begin{enumerate}
    \item We show that combining factual claims can yield a non-factual paragraph. 
    Existing factuality metrics have not correctly handled these non-factual generations.
    \item We collect 500 ambiguous names into a new dataset, AmbigBio, and each name corresponds to multiple entities in Wikipedia.
    \item We extend FActScore into D-FActScore, which can better assess non-factual generations stemming from entity ambiguity.
    \item We show that four open-source LLMs cannot properly handle entity ambiguity in the retrieved passages to generate a factual paragraph, yielding much lower D-FActScores compared to ChatGPT.
\end{enumerate}

\section{LLMs Combine Factual Claims into Non-factual Paragraph}
\label{section: Data preparation}
This section shows how to prompt LLMs to generate people biographies full of factual claims but combined in a non-factual way due to entity ambiguity.
We choose to generate people biographies for the following reasons:
(1) There are extensive prior works that use biography generation to evaluate the factuality of LLMs~\citep{min-etal-2023-factscore,asai2023self}.
(2) Given that name-based queries are common in online searches, it is likely that LLM users will directly request information about specific entities from LLMs.

\subsection{Collecting Names with Ambiguity}
\label{subsection: data}
To generate biographies with mixed information of multiple entities with the same name, we collect names with ambiguity from Wikipedia disambiguation pages using the \hyperlink{https://en.wikipedia.org/w/api.php}{Wikipedia API}.
A disambiguation page lists all the Wikipedia pages of the entities with the same name.
We randomly select 500 ambiguous names and call this collection \textit{AmbigBio} in our paper.
We use \textit{\textbf{target name}} to refer to a name in AmbigBio. 
On average, a target name corresponds to 4.6 entities in Wikipedia.

\subsection{Prompting LLMs to Generate Biographies}
We prompt LLMs to generate biographies with retrieval-augmented generation (RAG).

\textbf{Retrieval.}\;\;
Given a target name in AmbigBio, we retrieve the top-5 passages related to the name from Wikipedia using GTR~\citep{ni-etal-2022-large}.
The knowledge source for retrieval is the 2018-12-20 Wikipedia snapshot split into passages of 100 words~\citep{karpukhin-etal-2020-dense}.
The query used for retrieval is "\texttt{Tell me a bio of <name>}".
Some names in AmbigBio contain the string "\textit{(disambiguation)}", indicating that it is a disambiguation page.
The "\textit{(disambiguation)}" is removed when creating the query for retrieval.

\begin{table}[t]
\footnotesize
    \centering
    \begin{tabular}{p{22em}}
     \hline
     \texttt{Write an accurate, engaging, and concise biography of the person \textcolor{Bittersweet}{using only the provided search results} (some of which might be irrelevant) and cite them properly. (...)} \\
     \texttt{Document [1] (Title: ...) ...} \\
     \texttt{Document [2] (Title: ...) ...} \\
     \texttt{...} \\
     \texttt{Document [5] (Title: ...) ...} \\
     \texttt{Name of the person: <name>} \\
     \texttt{Answer: <answer>} \\
     \hline 
\end{tabular}
    \caption{The \textsc{Vanilla} prompt for prompting LLM.}
    \label{tab:prompt}
\end{table}

\textbf{Generation.}\;\;
We prompt LLMs with the top-5 retrieved passages to generate biographies for a target name with citations~\citep{nakano2021webgpt,gao-etal-2023-enabling}.
The LLM can only use the retrieved documents and must cite the retrieved passages in the output for attribution.
The reason to generate biographies with citations is to evaluate the attribution of the generated content.
In Appendix~\ref{subsection: citation recall}, we will show that even perfect citation attribution can still be non-factual.
We prompt the LLMs by \textsc{Vanilla} prompt (shown in Table~\ref{tab:prompt}) used in~\citet{gao-etal-2023-enabling} due to its superior performance in generating text with citations.
The title for each retrieved passage is included in the prompt.
The titles in Wikipedia sometimes contain parenthesis for disambiguation, e.g., \texttt{Dick Hanley (Swimmer)} in Figure~\ref{fig:illustration.pdf}.
We \textbf{do not} remove the words in the parenthesis when prompting the LLM to generate the biography, allowing the LLM to use this information for disambiguation.

We use 2-shot demonstrations to prompt the LLM.
The demonstration is similar to the prompt in Table~\ref{tab:prompt}, and the \texttt{<answer>} is replaced with a paragraph written by the authors.


\subsection{Large Language Models}
We use five LLMs with different sizes and alignment methods in our paper: Llama-13b-chat, Llama-70b-chat~\citep{touvron2023llama}, Vicuna-7b~\citep{vicuna2023}, Tulu-v2-13b-dpo~\citep{ivison2023camels}, and ChatGPT (\texttt{gpt-3.5-turbo-0301})~\citep{ChatGPT}.

\subsection{Categorizing LLM-Generated Paragraphs}
\label{subsection: Categorizing LLM-Generated Paragraphs}
We categorize the generated paragraphs based on the \textit{number of distinct entities} and the \textit{number of disambiguable biographies}, defined as follows.
The below definitions and categorization are better understood with the examples in Table~\ref{tab:output example}.

\textbf{Definition: (Named) Entity.}
A named entity is an object in real world that can be denoted with a proper noun.
In our paper, an entity is a real-world human with a corresponding Wikipedia page.

\textbf{Definition: (Entity) Mention.}
An entity mention is a specific instance when a named entity is referenced or mentioned within text.
An entity may be mentioned in different ways.

\textbf{Definition: Number of distinct entities in a paragraph.}
A paragraph can be decomposed into atomic facts following~\citet{min-etal-2023-factscore}.
Since the LLM is instructed to use only the retrieved documents to compose a paragraph, each atomic fact in the paragraph should originate from a retrieved passage, which is the Wikipedia page of an entity.
We attribute each atomic fact in the paragraph to a Wikipedia entity and collect the entities into a set; the number of distinct entities is the number of unique elements in the set.

\textbf{Definition: Number of distinguishable biographies in a paragraph.}
A paragraph might contain information about multiple entities, but its wording could mislead readers into believing it is about only one individual, making it seem like a single bio (as shown in Figure~\ref{fig:illustration.pdf}).
The number of distinguishable bios is given by the following operational definition: \textit{How many individuals with the target name will one identify in the paragraph without relying on prior or external knowledge?}
Note that a paragraph can contain more than one bio.

We categorize the LLM outputs into three types:

\begin{table*}[]
\footnotesize
    \centering
    \begin{tblr}{
      colspec = {X[m, 6em] X[m, 35em] X[m, 5em]},
      row{1} = {font=\bfseries}
    }
        \hline
        Type & LLM-generated paragraph &  Metrics \\
        \hline
        One bio one entity & \hlc[GreenYellow]{[Bio1]} \hlc[Apricot]{[Ent1]}  John Stewart, a Northern Irish politician born in 1983, has served as a Member of the Northern Ireland Assembly (MLA) for East Antrim since 2017. Stewart is a Unionist and previously worked in sales for Robinson's Shoemakers, a family business. Before joining politics, he attended Carrickfergus Grammar School and earned a Bachelor of Arts (BA) degree from Cardiff University. & \makecell{FS: 92\%    \\ D-FS: 92\%\\ Num bio: 1 \\ Num ent: 1} \\
        \hline
        One bio $N>1$ entities & \hlc[GreenYellow]{[Bio1]} \hlc[Apricot]{[Ent1]} John Hemming (born January 5, 1935) is a British historian and explorer who has made significant contributions to the study and understanding of the Incas and indigenous peoples of the Amazon Basin. Born in Vancouver to a British journalist mother, Alice Hemming OBE, and a father who had served in the First World War and wanted to ensure his son's birth in North America, John Hemming was raised in England from the age of two months. \hlc[Apricot]{[Ent2]} Hemming went on to study at Clapham Grammar School before \hlc[Apricot]{[Ent3]} earning a degree in Arithmetic and Theoretical, Atomic and Nuclear Physics at Magdalen College, Oxford. & \makecell{FS: 100\%    \\ D-FS: 71\%\\ Num bio: 1 \\ Num ent: 3} \\
        \hline
        $N>1$ bios $N>1$ entities & {\hlc[GreenYellow]{[Bio1]} \hlc[Apricot]{[Ent1]} Joseph F. Smith, the sixth President of The Church of Jesus Christ of Latter-day Saints, was born in 1838, and was the nephew of Joseph Smith, the founder of the Latter Day Saint movement. Additionally, Smith, who was the last president to have personally known the church’s founder, led the LDS Church. \hlc[GreenYellow]{[Bio2]} \hlc[Apricot]{[Ent2]} Joseph F. Smith, an American politician from Pennsylvania, was also born in 1920 and served in the United States House of Representatives. After a decorated military career, Smith was elected to represent Pennsylvania during the Ninety-seventh United States Congress.} & \makecell{FS: 87\%    \\ D-FS: 87\%\\ Num bio: 2 \\ Num ent: 2} \\
        \hline
    \end{tblr}
    \caption{Examples of three different types of LLM-generated paragraphs.
    \hlc[GreenYellow]{[Bio $i$]} denotes the start of the $i$-th distinguishable biography.
    \hlc[Apricot]{[Ent $i$]} denotes the subsequent information is about the $i$-th entity in Wikipedia that has the target name.
    We remove the citations ([1][2][3]) from the LLM-generated paragraphs here.}
    \label{tab:output example}
\end{table*}

\textbf{(1) One biography and one entity:}
The generated output only contains the information of one entity. 
In other words, all the information in the generated output points to the same Wikipedia entity whose name is the target name.

\textbf{(2) One biography and $N>1$ entities:}
The generated output mixes the information of different entities with the target name but does not provide sufficient disambiguation information.
A typical reader without prior knowledge will consider the whole paragraph a single bio of one individual.

\textbf{(3) $N>1$ biographies and $N>1$ entities:}
The paragraph contains information about multiple entities and provides enough context for readers without prior knowledge to disambiguate different entities in the paragraph.
Each biography describes one of the entities with the target name.


Ideally, the number of distinguishable bios should match the number of distinct entities.
When these two numbers agree, the LLM-generated paragraph provides enough disambiguation information in the paragraph and is likely factual.

\section{Existing Factual Precision Metrics}
We discuss why some factual precision metrics cannot properly assess the factuality of paragraphs with entity ambiguity, as shown in Figure~\ref{fig:illustration.pdf}.

\subsection{FActScore~\citep{min-etal-2023-factscore}}
\label{subsection: FActScore}
The key idea of FActScore is to decompose a long-form generation $y$ into a list of \textit{atomic facts} $\mathcal{A}_y$ - short sentences that convey one piece of information.
FActScore of a paragraph $y$ is defined by:
\begin{equation*}
\label{fs eqa}
    \text{\textsc{FS}}(y) = \frac{1}{|\mathcal{A}_y|} \sum_{a\in \mathcal{A}_y} \mathbbm{1}_{[a\text{ is supported by $\mathcal{C}$}]},
\end{equation*}
where $\mathcal{C}$ is a knowledge source for verifying the facts.
~\citet{min-etal-2023-factscore} use Wikipedia as $\mathcal{C}$ and prompt LLMs to generate people biographies as $y$.

\citet{min-etal-2023-factscore} show that FActScore can be obtained using an automatic pipeline: they instruct GPT3.5 (\texttt{text-davinci-003}) to decompose a long-form generation into atomic facts and use another $\text{LM}_{\text{EVAL}}$ to verify each fact.
They propose three different types of $\text{LM}_{\text{EVAL}}$ to verify an atomic fact:
(1) No-context LM: prompt an $\text{LM}_{\text{EVAL}}$ using '\texttt{<atomic fact> True or False?}'
(2) Retrieve $\to$ LM: retrieve $k$ passages from Wikipedia, construct the prompt by concatenating the retrieved passages, the atomic fact, and '\texttt{True or False?}', and prompt $\text{LM}_{\text{EVAL}}$ to answer.
(3) Nonparametric Probability (NP): an atomic fact is factual if the average token likelihood of the fact calculated using a nonparametric masked LM~\citep{min-etal-2023-nonparametric} exceeds a pre-defined threshold.

\textbf{Shortcoming of FActScore.}\;\;
FActScore is not designed to handle entity ambiguity because FActScore considers the factuality of each atomic fact independently without considering the whole paragraph.
We explain this using the example in Figure~\ref{fig:illustration.pdf}.
In the Retrieve $\to$ LM method, each atomic fact about Dick Hanley in the paragraph is verifiable by Wikipedia, leading to a 100\% FActScore. 
However, some facts are supported by \texttt{Dick Hanley (American Football)}, and others are supported by \texttt{Dick Hanley (Swimmer)}, indicating that there are two entities in the paragraph.
But this cannot be easily seen from the paragraph itself, and readers without any prior knowledge may believe that Dick Hanley is a swimmer who died in 1970, which is incorrect.
Since this paragraph does not allow readers to understand the information in a factual way, the paragraph should not be considered factual.

No-context LM and NP methods cannot properly handle entity ambiguity. 
For example, consider the fact "\textit{Dick Hanley passed away on December 16, 1970}".
This fact is factual if it is extracted from the biography of  \texttt{Dick Hanley (American Football)}.
Contrarily, if the fact is from the biography of \texttt{Dick Hanley (Swimmer)}, the fact is non-factual.
However, for the above atomic fact, no-context LM and NP methods always yield the same result and cannot consider whose biography the fact is extracted from.  
This makes them unable to distinguish entities with the same name.

\subsection{Citation Recall~\citep{gao-etal-2023-enabling}}
\label{subsection: citation recall}
Citation recall assesses the citation quality when generating text with citations by measuring whether the generated text is fully supported by the cited documents.
The core concept of citation recall is very similar to FActScore, and it suffers from the same problem as FActScore in the case of entity ambiguity.
We elaborate on this in Appendix~\ref{subsection: citation recall}.

\section{Disambig-FActScore (D-FActScore)}
\label{subsection: Contextual-FActScore}

We refine FActScore into D-FActScore to better address entity ambiguity in factuality evaluation. 
The definition of D-FActScore is presented in Section~\ref{subsection: D-FactScore definition}. 
Section~\ref{subsection: D-FActScore human eval} outlines the human annotation process for D-FActScore, with the outcomes of human evaluations detailed in Section~\ref{subsection: Human Evaluation Results}.

\subsection{Definition}
\label{subsection: D-FactScore definition}
We first define D-FActScore and then present the key concepts that motivate its definition.

\textbf{Definition.}\;
Let $y$ be a generated paragraph from an LLM.
Let $\mathcal{A}_{y}$ be a list of atomic facts decomposed from $y$.
We split $\mathcal{A}_{y}$ into $N$ disjoint \textit{fact groups} $\{ \mathcal{A}_{y,1},\cdots,\mathcal{A}_{y,N} \}$.
For two atomic facts $a, a' \in \mathcal{A}_{y}$, they will be grouped into the same fact group $\mathcal{A}_{y,i}$ if a reader without prior or external knowledge may think that $a$ and $a'$ are about the same \textit{individual} when reading the paragraph.
We use the term \textit{individual} to refer to a character perceived by the reader; this is different from an \textit{entity} that exists in the real world.
For example, all the atomic facts in Figure~\ref{fig:illustration.pdf} are grouped into a single fact group since the paragraph looks like it is about the same individual named Dick Hanley.

For each group of atomic facts $\mathcal{A}_{y,i}$, we use entity linking to find an entity $e_i^{*}$ in the knowledge source $\mathcal{C}$ that best matches the facts in $\mathcal{A}_{y,i}$.
Let the subset of the knowledge source related to $e_i^{*}$ denoted by $C_i^{*}$.
For all atomic facts $a\in\mathcal{A}_{y,i}$, they will be verified using $C_i^{*}$ instead of using the whole $C$.
The D-FActScore of $y$ is defined as follows:

\begin{equation*}
\begin{gathered}
  \text{D-FS}(y) = \frac{1}{|\mathcal{A}_y|} \sum_{\mathcal{A}_{y_i} \in \mathcal{A}_{y}} \sum_{a \in \mathcal{A}_{y_i}} \mathbbm{1}_{[a \text{ is supported by } C_i^*]},
\end{gathered}
\end{equation*}
In our paper, $y$ is a paragraph generated in Section~\ref{section: Data preparation}, $\mathcal{C}$ is Wikipedia, and $C_i^{*}$ is the Wikipedia page of $e_i^{*}$.

D-FActScore differs from FActScore in two aspects.
\textbf{Difference 1:} D-FActScore groups atomic facts by their originating paragraph before verifying them.
\textbf{Difference 2:} D-FActScore restricts that all the facts in the same group must be verified using the information of the same entity.

\textbf{Motivation.}\;
Atomic facts from a paragraph often have connections, so evaluating their factuality should consider these relationships, rather than verifying each fact independently as FActScore does. 
Grouping atomic facts helps manage these connections: Facts within the same group \textit{look like} they are about the same individual to the readers, so their truthfulness should be confirmed using the same entity in the knowledge source.
For example, the atomic facts in Figure~\ref{fig:illustration.pdf} belong to one group, and they can only be supported by either the Wikipedia page of \texttt{Dick Hanley (Swimmer)} or \texttt{Dick Hanley (American Football)}, but not both.

By organizing facts into groups and limiting the source of fact verification for each group, D-FActScore assesses the factuality of a paragraph with entity ambiguity more accurately. 
These two key differences mark the most significant difference between D-FActScore and FActScore, enabling D-FActScore to handle entity ambiguity.

\subsection{Annotating D-FActScore by Humans}
\label{subsection: D-FActScore human eval}
We conduct human evaluation to calculate the D-FActScore of paragraphs generated in Section~\ref{section: Data preparation}.
The annotators are presented with a paragraph, the atomic facts decomposed from the paragraph, and all the Wikipedia pages of the entities with the target name.
The exact instructions, annotation interface, and agreement rate between annotators are shown in Appendix~\ref{section: Human Evaluation}.
The annotations are conducted via the following steps.

\textbf{Step 1: Decompose atomic facts.} 
Following~\citet{min-etal-2023-factscore}, we use GPT-3.5 to extract atomic facts from a paragraph.

\textbf{Step 2: Determine the number of bios.} 
We instruct the annotators to determine the number of biographies based on the paragraph, neglecting any prior knowledge or the Wikipedia pages we prepare.
Identifying more than one biography indicates that the paragraph provides enough information to separate the biographies of distinct individuals.
\textbf{This step essentially splits atomic facts into groups} since atomic facts of the same biography are about the same individual, so they fall into the same group.
We also ask the annotators to link each biography to an entity's Wikipedia page we present.
Even though determining the number of biographies in a paragraph is based on personal interpretation, we find that annotators reach a high level of agreement on the number of bios.

\textbf{Step 3: Verifying atomic facts.}
For each atomic fact, the annotators first check if the fact is \texttt{Irrelevant} to the target name.
If it is not \texttt{Irrelevant}, verify whether the atomic fact is \texttt{Supported} or \texttt{Not-Supported} using one entity's Wikipedia page.
Recall that the atomic facts in the same group belong to one of the biographies in the original paragraph and the biography is linked to an entity's Wikipedia page in Step 2.
All the atomic facts in the same group are verified using the same linked entity's Wikipedia page.

We calculate the D-FActScore based on the human annotation results.
Three annotators from Upwork are hired to label 300 paragraphs generated in Section~\ref{section: Data preparation} from Llama-13b-chat, Tulu-v2-13b-dpo, and ChatGPT; 100 paragraphs per model.
We choose these models since they include open-source and proprietary models and have different alignment training methods.


\subsection{Human Evaluation Results}
\label{subsection: Human Evaluation Results}
Aside from the annotation of D-FActScore elaborated above, we ask the annotators to perform additional annotation to allow us to calculate the number of distinct entities and the FActScore of the paragraph.
FActScore is annotated based on the definition in~\citet{min-etal-2023-factscore}.
The number of distinct entities is calculated based on the definition in Section~\ref{subsection: Categorizing LLM-Generated Paragraphs}.
Details are in Appendix~\ref{appendix: Human Evaluation of FActScore}.
The human annotation results of D-FActScore and FActScore are presented in Table~\ref{tab:human_eval}. 
We have the following observations.

\begin{table}[t]
    \centering
    \begin{subtable}{\columnwidth}
        \centering
        \begin{tabular}{ccccc}
        \hline
         \textbf{Model}&  \textbf{FS}&  \textbf{D-FS}&  \textbf{\# bio}& \textbf{\# ent.}\\
         \hline
         ChatGPT&  98.3 &  92.1&  2.2 & 2.3 \\
         chat-13b&  94.8&  78.4&  1.0 & 1.7\\
         Tulu&  91.9&  83.2&  1.3 & 1.7\\
         
         \hline
        \end{tabular}
        \caption{Human evaluation}
    \label{tab:human_eval}
    \end{subtable}
    \begin{subtable}{\columnwidth}
        \centering
        \begin{tabular}{ccccc}
        \hline
         \textbf{Model}&  \textbf{FS}&  \textbf{D-FS}&  \textbf{\# bio}& \textbf{\# ent.}\\
         \hline
         ChatGPT&  98.7&  96.3&  2.2& 2.3\\
         chat-13b&  95.3&  86.4&  1.1& 1.5\\
         Tulu&  95.8&  88.5&  1.3& 1.7\\
         
         \hline
        \end{tabular}
        \caption{Automatic evaluation}
    \label{tab:human_eval_set_automatic}
    \end{subtable}
    \caption{Human and automatic evaluations of FActScore and D-FActScore conducted on the same set of paragraphs.
    FS: FActScore, D-FS: D-FActScore, \# bio: average number of separable biographies in one paragraph, \# ent.: average number of distinct entities in one paragraph, chat-13b: Llama-13b-chat.}
    \label{table: subset result}
\end{table}

\paragraph{FActScore overestimates the factuality of the LLM-generated paragraphs.}
This is because all models mix the facts of multiple entities in a single biography, and FActScore considers these misleading paragraphs factual as long as each atomic fact can be supported by Wikipedia.
D-FActScore does not have this problem by construction.
The gap between FActScore and D-FActScore can be interpreted as the tendency of an LLM to mix the facts of multiple entities in a non-factual way.

\paragraph{FActScore and D-FActScore yield different model rankings.}
The FActScore of Llama-13b-chat is higher than Tulu-v2-13b-dpo by 2.9\%, but D-FActScore reveals that Llama-13b is less factual than Tulu by 4.8\%.
Going through the paragraphs generated by Llama-13b and Tulu, we find that Llama-13b is good at copying sentences from the retrieved passages to form a paragraph, thus having a higher FActScore.
On the other hand, Tulu does not always copy the retrieved content but is better at disambiguating entities in the retrieved passages than Llama-13b, yielding a higher D-FActScore.

\paragraph{ChatGPT can utilize information in the retrieved passages to disambiguate entities.}
ChatGPT has the highest D-FActScore, and the average number of biographies and entities is almost the same.
This implies that ChatGPT can distinguish entities with the same name in the retrieved passages and generate a factual paragraph that provides the readers with sufficient information to disambiguate those entities.

\section{Automatic Evaluation of D-FActScore}
\label{main section: Automatic Evaluation of D-FActScore}
Human evaluation is time-consuming and expensive.
Hence, we devise an automatic pipeline to estimate D-FActScore (Section~\ref{subsection: D-FActScore automatic eval}) and show that it can approximate the D-FActScore obtained by human annotation (Section~\ref{subsection: Human VS Automatic Evaluation}).
We then use the automatic pipeline to evaluate the generation from five LLMs (Section~\ref{subsection: Comparing Different LLMs and Different Demonstrations}).

\subsection{Automatic Evaluation}
\label{subsection: D-FActScore automatic eval}
The automatic evaluation of D-FActScore resembles that of human annotations.

\textbf{Step 1: Decompose atomic facts from a paragraph using GPT3.5.}

\textbf{Step 2: Split facts into fact groups.}
We give GPT3.5 the LLM-generated paragraph and the atomic facts decomposed in Step 1, and we ask GPT-3.5 to split the atomic facts into groups, where each group corresponds to the atomic facts of one distinguishable biography in the paragraph.
GPT3.5 is instructed to use the paragraph only to group the facts for distinct biographies.
We use 4-shot demonstrations in this step.

\textbf{Step 3: Verifying atomic facts.}
We verify the factuality of facts in the same fact group using the Wikipedia page of the same entity; a fact group corresponds to one biography in the original paragraph.
For each biography and its corresponding fact group, we perform entity linking to find a Wikipedia entity that best matches the individual of the biography and verify the facts in that bio based on the Wikipedia page of that entity.
Precisely, we use Retrieve $\to$ LM in~\citet{min-etal-2023-factscore} to prompt ChatGPT to verify the atomic fact based on the Wikipedia of that entity.
Details of entity linking and the prompts used in automatic evaluation are elaborated in Appendix~\ref{subsection: entity linking}.

\subsection{Human VS Automatic Evaluation}
\label{subsection: Human VS Automatic Evaluation}

We compare the result of D-FActScore obtained with human and automatic evaluation in Table~\ref{table: subset result}.
We have the following findings.

\textbf{Automatic and human evaluation of D-FActScore shows the same model ranking.}\;\;
We find the factuality ranking among the three models based on D-FActScore in Table~\ref{tab:human_eval_set_automatic} agrees with the ranking in Table~\ref{tab:human_eval}:
ChatGPT is the most factual, and Llama-13b-chat is the least factual.
This shows that the automatic evaluation of D-FActScore provides a reliable estimation of the relative factuality of different LLMs.

\textbf{D-FActScore obtained with automatic evaluation is higher than human evaluation.}\;\;
Compared with the human evaluation result of Table~\ref{tab:human_eval}, D-FActScore obtained using automatic evaluation is higher, and the absolute error is at most 8\%.
This observation also aligns with~\citet{min-etal-2023-factscore}, which shows that using Retrieve $\to$ LM to estimate FActScore can yield a higher FActScore compared to human evaluation.

\textbf{Automatic evaluation can determine the number of biographies accurately.}\;\;
The correctness of D-FActScore's automatic evaluation strongly depends on the second step: splitting atomic facts into groups corresponding to biographies of different individuals.
The number of groups is the number of distinguishable biographies.
By comparing the number of bios obtained using automatic and human evaluation in Table~\ref{table: subset result}, we find that GPT3.5 can accurately determine how many biographies there are in a paragraph.
The difference between the number of biographies obtained by human evaluation and automatic evaluation differs within 0.1 in all three models, justifying using GPT3.5 to split the atomic facts into fact groups.

\textbf{Per-paragraph D-FActScore of human and automatic evaluation results show a high correlation.}\;\;
We calculate the per-paragraph Pearson correlation coefficient $r$ between D-FActScore obtained with human evaluation and automatic evaluation, and we also calculate $r$ between the number of bios obtained by human and automatic evaluation. 
The results shown in Table~\ref{tab: correlation coefficient} in the Appendix reveal that the Person’s $r$ of D-FActScore between human and automatic evaluation is moderate (ChatGPT) to strong (Llama and Tulu).  
If we consider the D-FActScore of the paragraphs generated by all three models together and calculate Pearson's r, the correlation coefficient is 0.75, which is a strong correlation. 
We also find that for all three models, the number of bios obtained using human and automatic evaluation are all very strongly correlated (Pearson's $r\geq0.8$).
The above experiment results further prove the effectiveness of the automatic evaluation pipeline.

\subsection{Different LLMs and Demonstrations}
\label{subsection: Comparing Different LLMs and Different Demonstrations}
After showing that D-FActScore can be estimated using automatic evaluation, we use automatic evaluation to evaluate the D-FActScores of paragraphs generated using all 500 names in AmbigBio by five LLMs and two types of demonstrations: with and without name ambiguity.

\textbf{(1) With name ambiguity}:
The names in the demonstrations are ambiguous names in Wikipedia; the retrieved passages include Wikipedia pages of different entities with the same name.
This is the default setting in the previous experiments.

\textbf{(2) Without name ambiguity}:
The name in the demonstration is unambiguous (there is no disambiguation page for that name in Wikipedia).
The retrieved results contain the passages from that entity's Wikipedia page and possibly some unrelated passages.

The \texttt{<answer>} in the demonstrations with and without name ambiguity only contains the information of one entity with the target name.
We have the following findings.

\textbf{Demonstrations with name ambiguity do not make the outputs more factual.}\;\;
In Table~\ref{tab: automatic eval full}, we compare the results of demonstrations with and without ambiguity.
In the \textit{with ambiguity} demonstration, the demonstration \texttt{<answer>} only contains a biography that includes the information of one entity, and we hope the LLM can know how to handle target names with ambiguity better.
However, we do not see a higher D-FActScore when using demonstrations with name ambiguity.

\begin{table}
\footnotesize
    \centering
    \begin{tabular}{cccccc}
         \hline
         & \textbf{FS} & \textbf{D-FS} & \textbf{\# bios} & \textbf{\# ent.} \\
         & \multicolumn{4}{c}{\textit{with name ambiguity / without name ambiguity}} \\
         \hline
         ChatGPT & 96.7 / 96.7 & 95.2 / 94.3 & 2.3 / 2.1 & 2.3 / 2.3 \\
         chat-13b & 94.6 / 94.3 & 86.0 / 83.2 & 1.1 / 1.1 & 1.6 / 1.8 \\
         chat-70b & 94.8 / 94.0  & 86.4 / 85.6 &  1.6 / 1.8 & 2.1 / 2.3 \\
         Tulu & 94.2 / 95.2 & 88.5 / 90.2 & 1.4 / 1.4 & 1.8 / 1.7 \\
         Vicuna & 90.0 / 93.4 & 87.7 / 88.9 & 1.3 / 1.3 & 1.6 / 1.7 \\
         \hline
    \end{tabular}
    \caption{The results of automatic evaluation on 500 passages generated with 500 names in AmbigBio for each model.
    We report the result when the demonstration includes examples \textit{with name ambiguity} and \textit{without name ambiguity} on the left and right of each cell.
    The abbreviations are the same as Table~\ref{table: subset result}.}
    \label{tab: automatic eval full}
\end{table}

\textbf{Open-source LLMs lag behind ChatGPT.}\;\;
All the open-source models we use have a D-FActScore much lower than that of ChatGPT.
Furthermore, the entity per paragraph is higher than biography per paragraph for open-source models, showing that the open-source models cannot distinguish entities with name ambiguity and generate a factual paragraph.
This highlights a potential direction of improvement for open-source LLMs.

\textbf{Scaling the model size does not improve D-FActScore too much.}\;\;
The D-FActScore of Llama-70b-chat is only higher by Llama-13b-chat by less than 2\%, which is a marginal improvement considering the disproportional size.
The main difference between the paragraph generated by Llama-13b-chat and 70b variant is that Llama-70b-chat tends to include facts about more entities in the paragraph, but it still does not properly disambiguate the entities in the paragraph, making merely no improvement to D-FActScore.

\textbf{ChatGPT fully uses the retrieved documents.}\;\;
By examining the passages retrieved by GTR, we estimate that the top-5 passages provided to the LLM contain 2.2 distinct entities with the target name on average.
Meanwhile, the passages generated by ChatGPT also contain around 2.1 to 2.3 biographies and distinct entities on average.
This indicates that ChatGPT can include diverse information presented in the retrieved passages.
This is a desirable behavior since the answer to an ambiguous question should cover the answer to as many disambiguated questions~\cite {min-etal-2020-ambigqa,stelmakh-etal-2022-asqa}.

\textbf{The central figure of the biography is not always the most common Wikipedia entity.}\;\;
When multiple entities with the target name are included in the top-5 retrieved passages, the LLM can generate a biography for any of them.
We want to answer the question: 
When there is only one biography and one entity in the generated paragraph, but multiple valid entities exist in the retrieved passages, is the central figure of the biography always the most common entity among the retrieved entities?
We hypothesize that popular entities are more likely to be included in the LLM's training data more times, making LLM more familiar with those common entities and prone to use them as the central figure.
We assess how common an entity is based on the page view of their Wikipedia over the past year.
However, for all LLMs, we find that in only 45\% to 55\% of the cases, the LLM picks the most popular entity to write a biography for.
This shows that an entity's popularity does not strongly affect which entity the LLM picks to generate a biography for.

\section{Related Work}
\label{section: Related Work}
\paragraph{Factuality Evaluation}
Evaluating the factuality of texts is an active subfield in NLP.
Some prior works formulate factuality evaluation as an uncertainty estimation problem and consider generation that models are less confident to be less factual~\citep{liu-etal-2022-token,zhang-etal-2023-enhancing-uncertainty, manakul-etal-2023-selfcheckgpt}.
Many recent works focus on evaluating the precision of the citation when generating texts with citations and use QA or NLI models to determine if the statements in the generation can be supported by the cited documents~\citep{rashkin2023measuring,liu-etal-2023-evaluating,gao-etal-2023-enabling, yue-etal-2023-automatic}.
Our work is largely based on FActScore~\citep{min-etal-2023-factscore}, and we improve FActScore by resolving its inability to handle entity ambiguity.

\paragraph{Evaluating Faithfulness in Summarization}
Our work is somewhat related to multi-document summarization since generating a biography based on the retrieved documents is like summarizing the contents into a biography; the main difference is that the retrieved documents can be irrelevant to the biography to be generated.
The concept of \textit{faithfulness} in summarization, whether the summaries are factually consistent with the source documents, is quite similar to factual precision.
Many prior works focus on benchmarking and improving the automatic and human evaluation of faithfulness~\citep{kryscinski-etal-2020-evaluating, pagnoni-etal-2021-understanding, laban-etal-2022-summac, krishna-etal-2023-longeval}.

Unfaithfulness due to incorrect or incomplete coreference is a well-known problem in summarization~\citep{pagnoni-etal-2021-understanding,zhang-etal-2023-extractive}.
The problem of combining facts into non-factual paragraphs identified in our paper is related to incorrect coreference but is different since incorrect coreference stems from ambiguous anaphors instead of ambiguous entities. 
Unfaithfulness due to coreference ambiguity is very hard to evaluate using automatic evaluation metrics, as automatic evaluation metrics have low correlations with human evaluation results~\cite {pagnoni-etal-2021-understanding}.
D-FActScore, which can be estimated by automatic evaluation, can properly evaluate the non-factual contents stemming from entity ambiguity.

\paragraph{Ambiguous Question Answering}
Our work uses an ambiguous name to prompt the LLM to generate a biography.
This is highly related to ambiguous question answering~\citep{min-etal-2020-ambigqa,stelmakh-etal-2022-asqa}, where a question can have multiple answers based on how the question is interpreted.
\citet{min-etal-2020-ambigqa} estimates that about 23\% of the questions in \textsc{AmbigNQ} dataset are due to entity ambiguity, which is the focus of our work.
Most prior works answer an open-domain ambiguous question by first generating disambiguated questions and generating an answer for each disambiguated questions~\citep{gao-etal-2021-answering,min-etal-2021-joint,shao-huang-2022-answering,kim-etal-2023-tree}.
Some works do not disambiguate the ambiguous question and generate the answer directly~\citep{stelmakh-etal-2022-asqa,gao-etal-2023-enabling}.
We do not perform the disambiguation step because we want to know if the LLMs can perform well when we do not explicitly disambiguate the target entity.

Evaluation metrics of ambiguous QA are mostly based on string matching, where the model-generated answers are compared with the ground truth answer, which contains the ground truth answer to each disambiguated question~\citep{min-etal-2020-ambigqa,stelmakh-etal-2022-asqa}.
These existing metrics are not suitable for evaluating the paragraphs generated in Section~\ref{section: Data preparation} since we do not have a ground truth biography for each entity.
Additionally, metrics based on string matching may not properly consider the case when answers to disambiguated questions are combined misleadingly.

\section{Conclusion and Discussion}
We show that combining factual information yields a non-factual paragraph due to entity ambiguity, and this kind of paragraph can be easily generated by LLMs using RAG.
We further reveal that current factuality metrics cannot correctly assess the factuality of these non-factual paragraphs.
To resolve the evaluation problem, we modify FActScore into D-FActScore by focusing on entity disambiguation.
We conduct human evaluations of D-FActScore to compare the factuality of different LLMs and show that D-FActScore of open-source LLMs lag behind ChatGPT, indicating that open-source LLMs cannot handle ambiguous entities in the retrieved documents to form a factual narrative.
We propose a pipeline for automatic evaluation of D-FActScore and show it aligns with human evaluation results to a reasonable extent.
We encourage future researchers to use AmbigBio and evaluate the factuality of generated paragraphs using D-FActScore.
The ambiguous name collection AmbigBio and the codes for D-FActScore are available at \url{https://github.com/d223302/Merging-Facts-Crafting-Fallacies}.


Our findings underscore LLMs' difficulties in generating accurate content when retrieving from Wikipedia with entity ambiguity.
The scenario in our paper is simplified, given that Wikipedia is not the sole information source about entities. 
Even using high-quality contents from Wikipedia, open-source models struggle to differentiate between entities accurately. 
In more realistic situations where LLMs draw from a broader and more nuanced content pool on the Web, and the named entities used to query LLMs may not appear in the training data, distinguishing between different entities becomes even more challenging. 
Overcoming this issue is vital for the reliability of LLMs with RAG.

\section*{Limitations}
We see the following limitations in our work.

\paragraph{The content we evaluate}
In our study, we focus on evaluating D-FActScore of human biographies. 
We justify the reasons for doing this in Section~\ref{section: Data preparation}.
In fact, the difficulty of factuality evaluation due to entity ambiguity can happen in more diverse contents, and the core concepts and evaluation procedure of D-FActScore are general and can be applied to more diverse contents. 

\paragraph{Using Wikipedia for RAG and fact-checking}
We use Wikipedia for RAG and fact-checking. 
This is a very common protocol in the literature of RAG research since Wikipedia is the most commonly used knowledge source. 
For example, Self-RAG~\citep{asai2023self} also retrieves documents from Wikipedia to generate biographies and uses Wikipedia to evaluate the factuality of the generated biographies. 
It is important to explore the case when the RAG does not use Wikipedia for generation, and we will leave extending the data to other sources as future works.

\paragraph{Fact-checking beyond Wikipedia}
In our paper, we evaluate D-FActScore by relying on Wikipedia of the source for fact-checking.
This is a common protocol in fact-checking, which uses a structured knowledge source like Wikipedia or other databases to verify the facts.
In the case of verifying facts using unstructured knowledge sources, more effort may be required to disambiguate entities in the knowledge source.
To properly calculate D-FActScore, we need to be able to split the information in the knowledge source into the information of distinct entities.
That is, we need to disambiguate the articles in the knowledge source, for example, verifying the biography of a CS PhD student who is not in Wikipedia and using the contents on the Web for verification.
Let's say we find a LinkedIn page and some semantic scholar pages with the same name as our target entity.
To disambiguate the entities, we need to match semantic scholar pages with LinkedIn profiles. 
This can be done by various methods, including but not limited to (1) checking if the LinkedIn profile already is linked to a semantic scholar page, or (2) checking if the LinkedIn profile is linked to the person's homepage and checking if one can find a semantic scholar link on that page. 
This can all be done automatically with tool-augmented and retrieval-augmented LLMs.
However, the above process may be challenging when the knowledge source is more complicated and provides too little information for disambiguation.
We leave verifying facts using the contents on the Web in future work.

\paragraph{Beyond entity ambiguity}
Our paper focuses on using entity ambiguity to create passages full of factual claims but are overall non-factual.
However, entity ambiguity is not the only reason that may make factual claims be combined to form a non-factual narrative.
For example, consider the following two factual atomic facts: (1) \textit{Mountain Fuji is the highest mountain in Japan}, and (2) \textit{the population of Tokyo is about 14M}.
The following sentence composed with the two atomic facts is obviously nonsensical: \textit{Because Mountain Fuji is the highest mountain in Japan, the population of Tokyo is about 14M}.
In this sentence, the atomic facts are factual when they are considered independently, but the causal relation between these two facts is non-factual.
We do not consider/evaluate this kind of non-factuality, nor do prior works.
We leave this topic in future work.

\paragraph{Using GPT3.5 to extract atomic facts}
In our paper, we rely on using GPT3.5 (\texttt{text-davinci-003}) to extract atomic facts and split atomic facts into groups (Section~\ref{main section: Automatic Evaluation of D-FActScore}).
The reason for using GPT3.5 is to match the experiment setup of~\citet{min-etal-2023-factscore}, which also relies on GPT3.5 to extract the atomic facts.
Using the same model to extract atomic facts makes it easier for us to compare with them.
However, GPT3.5 was recently deprecated by OpenAI, making it impossible to reproduce the results of~\citet{min-etal-2023-factscore} and the experiment in our paper.
To alleviate this issue, we show in Table~\ref{tab:human_eval_set_automatic_chatgpt appendix} in Appendix~\ref{appendix: Using ChatGPT to Extract and Group Atomic Facts} that using ChatGPT (\texttt{gpt-3.5-turbo}) to extract the atomic facts and split the atomic facts into groups yields almost the same result as using GPT3.5.
All our observations in Section~\ref{main section: Automatic Evaluation of D-FActScore} hold when using ChatGPT to extract and group the atomic facts.

We do not see specific risks or harm in our paper.

\section*{Acknowledgements}
We thank the reviewers for providing detailed feedback and actionable suggestions, which helped us strengthen our paper.
We also thank the senior committee members for monitoring the reviewing process.
We thank Sewon Min for providing valuable, actionable, concrete feedback and insights on the paper.
Cheng-Han Chiang is supported by a Google PhD Fellowship and a Ph.D. scholarship program by Delta Electronics.
We thank the National Center for High-performance Computing (NCHC) of National Applied Research Laboratories (NARLabs) in Taiwan for providing computational and storage resources.

\bibliography{custom}

\appendix

\section{Generating Paragraphs from LLMs}
For all open-source models, we use a temperature of $1$ and top-p sampling with $p=0.95$.

\section{Human Evaluation}
\label{section: Human Evaluation}
We hire three freelancers with experience in fact-checking from Upwork to annotate the D-FActScore in Section~\ref{section: Human Evaluation}.
We have 300 paragraphs from 3 models to annotate.
We follow~\citet{min-etal-2023-factscore} to assign two annotators to label 10\% of the paragraph (10 paragraphs for each LLM) to calculate the agreement rate, and the remaining 90\% of the paragraphs are annotated by one annotator.
We evenly distribute the paragraphs generated by different LLMs to the annotators, so each annotator labels 30, 40, and 40 paragraphs for each model.
Each paragraph has, on average, 21 atomic facts to label \texttt{Supported}, \texttt{Not-Supported}, and \texttt{Irrelevant}.

\begin{figure*}[t!]
    \centering
    \includegraphics[width=1.0\linewidth]{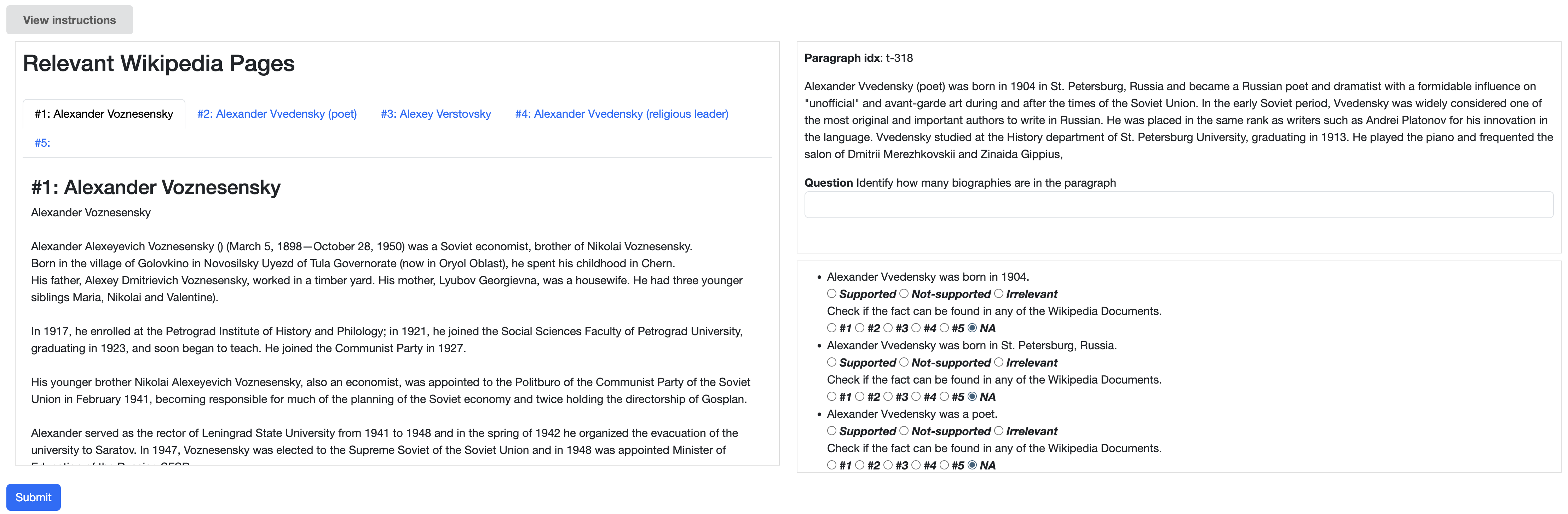}
    \caption{The interface used for annotation.}
    \label{fig:interface.png}
\end{figure*}

\begin{figure*}[t!]
    \centering
    \includegraphics[width=1.0\linewidth]{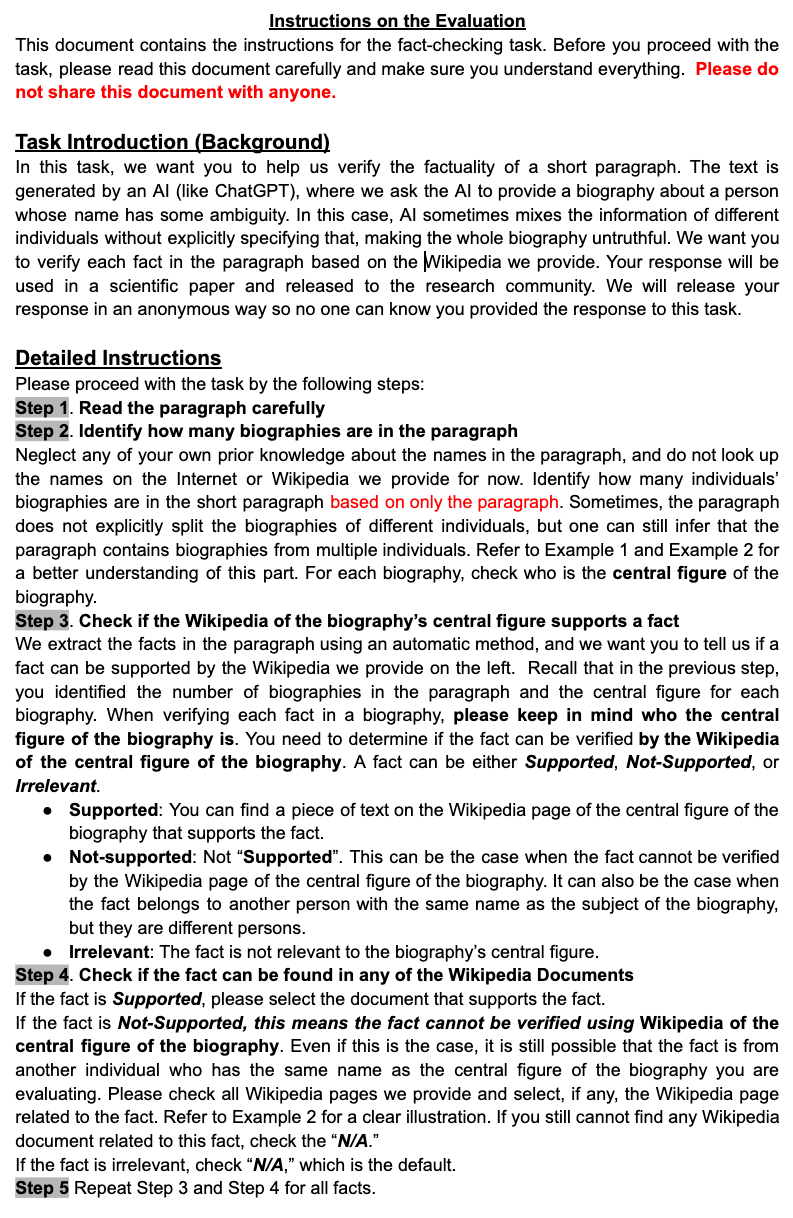}
    \caption{The instructions used for annotation.
    We do not show the examples in the instructions in this figure.}
    \label{fig:instructions.png}
\end{figure*}

We use Amazon Mturk Sandbox as the annotation platform, and the annotation interface is shown in Figure~\ref{fig:interface.png}.
The instructions are shown in Figure~\ref{fig:instructions.png}, and we include two example annotations.
After the annotators read the instructions, they will be given two simplified testing examples to test their understanding of the instructions.
If the annotation results on the testing examples do not match our expectation\footnote{While D-FActScore has some subjectivity due to the narrative of the paragraph, whether a fact is factual according to Wikipedia is mostly undebatable.}, we will discuss with the annotators with the results and clarify their understanding of the instructions.

We find that the agreement rates between annotators are quite high.
We calculate the agreement rate by the percentage when two annotators give the same result on the annotation.
The agreement rates on the number of bios for Llama-13b-chat, Tulu-v2-13b-dpo, and ChatGPT are 90\%, 90\%, and 100\%, respectively.
The agreement rates on whether an atomic fact is \texttt{Supported}, \texttt{Not-Supported}, or \texttt{Irrelevant} for Llama-13b-chat, Tulu-v2-13b-dpo, and ChatGPT are 74.4\%, 85.3\%, and 84.2\%, respectively. 

The annotators are informed about the purpose of the data collection and are aware that the data collected will be shared with the research community.
The annotators take, on average, 3 to 5 minutes to complete the annotation of one paragraph, and we pay them 2 USD for annotating one paragraph.
This leads to an hourly wage of about 24-40 USD. 

\subsection{Human Evaluation of FActScore}
\label{appendix: Human Evaluation of FActScore}
Based on the definition of FActScore in~\citet{min-etal-2023-factscore}, each atomic fact is verified using the whole knowledge source $\mathcal{C}$.
Following this definition, we ask human annotators to check if each atomic fact can be supported by Wikipedia.
Unlike the annotation of D-FActScore, we do not ask the annotators to consider entity ambiguity or the context where the atomic fact is extracted.
We also do not restrict the annotators to verify the atomic fact using a subset of the Wikipedia of the linked entity.
Instead, as long as an atomic fact can be verified with any Wikipedia page, the atomic fact should be considered \texttt{Supported}.

Our paper suggests that FActScore, while not initially tailored for paragraphs featuring entity ambiguity, may not fully capture the nuances in the evaluations of paragraphs outlined in Section~\ref{section: Data preparation}. Our contribution seeks to identify and address this area for improvement within the original framework of FActScore, as described in~\citet{min-etal-2023-factscore}, with the goal of refining the metric to more adeptly handle diverse content scenarios.

\subsection{Human Evaluation of Number of Distinct Entities}
\label{appendix: Human Evaluation of Number of Distinct Entities}
To evaluate the number of distinct entities by human evaluation, we give the human annotators the atomic facts decomposed from the paragraph.
For each fact, we ask the annotators to use the Wikipedia pages we provide, which consists of the Wikipedia pages of the entities with the target name, to determine which entity the given atomic fact refers to.
In this step, the annotators are instructed \textbf{not to} consider the context of the atomic fact, but consider the atomic fact as a standalone fact and find which entity this standalone atomic fact refers to.
After we obtain the entity of each atomic fact in the paragraph, we can count the number of unique entities for the atomic facts in the paragraph; this number is the number of distinct entities.

\section{Automatic Evaluation}
\label{section: automatic evaluation}
The automatic evaluation of D-FActScore comprises of three steps:
(1) Split the biography into atomic facts.
We follow~\citet{min-etal-2023-factscore} and use \texttt{text-davinci-003} to split the paragraph into atomic facts using in-context learning.

(2) Split the atomic facts into groups, where each group of atomic facts corresponds to one distinguishable biography.
We also use \texttt{text-davinci-003} in this step, where we provide 4-shot demonstrations to teach the LLM how to split the atomic facts into groups.
Two of the four demonstrations contain multiple distinguishable biographies, and the atomic facts from each biography should be grouped together. 
The other two demonstrations do not contain multiple biographies, and all the atomic facts in the paragraph belong to the same group.
We ask the LLM to insert "\texttt{===}" among atomic facts that belong to different groups of atomic facts.
Two demonstrations are shown in Table~\ref{tab:split bio prompt}.

(3) Verify each group of atomic facts, which we detailed in Appendix~\ref{subsection: entity linking}.

\subsection{Entity Linking}
\label{subsection: entity linking}
After we split the atomic facts into groups of atomic facts, we need to determine the entity each group refers to.
Recall that each group of atomic facts corresponds to a biography from the long-form generation, so the goal of this step is to assign a Wikipedia entity for this group of facts and use the Wikipedia page of that entity to verify all the atomic facts in this group.

For each group of atomic fact $\mathcal{A}_{y,i}$, we find its corresponding entity $e_i^{*}$ by iterating over all possible entities in Wikipedia and find the entity that maximally supports the facts in $\mathcal{A}_{y,i}$.
This can be expressed by
\begin{equation}
\label{eq: entity linking}
    e_i^{*} = \arg\max_{e_k\in\mathcal{C}}{\frac{1}{|\mathcal{A}_{y,i}|}\sum_{a\in\mathcal{A}_{y,i}} \mathbbm{1}[a\text{ is supported by $e_k$}]},
\end{equation}
where $\mathcal{C}$ is the whole Wikipedia.
However, calculating Equation~\ref{eq: entity linking} requires iterating over all the entities in Wikipedia, which is infeasible.
Thus, we approximate Equation~\ref{eq: entity linking} by replacing the $\arg\max$ over $\mathcal{C}$ with $\arg\max$ over all the entities in the top-5 passages retrieved with GTR. 
(Recall that the paragraphs are generated based on the top-5 passages retrieved using GTR)

After this process, we obtain $\{ e_1^*, \cdots, e_N^* \}$.
In very few cases, the optimal entity $e_i^*$ and $e_j^*$ for two groups of atomic facts $\mathcal{A}_{y,i}$, $\mathcal{A}_{y,j}$ might be the same, and we use Hungarian algorithm to assign the entity to each group of atomic facts by maximizing the overall D-FActScore.
However, we find that the results of using the Hungarian algorithm are quite similar to the results of not using the Hungarian algorithm, so we remove the Hungarian algorithm in our final implementation for a shorter run time.

\subsection{Per-Paragraph Correlation Coefficient between Human and Automatic Evaluations}
Table~\ref{tab: correlation coefficient} shows the per-paragraph Pearson correlation coefficient $r$ between the D-FActScore and the number of bios obtained by automatic evaluation and human evaluation.
The reason that Pearson's r of ChatGPT is not as strong as the other two models is mainly because the D-FActScores of ChatGPT are mostly distributed in a smaller range (from 90 to 100) due to its high D-FActScores, and a slight variation of the evaluation result may change the Pearson's r significantly.
\begin{table}[h]
  \centering
  \begin{tabular}{c|cc}
    \hline
    \textbf{Model} & \textbf{D-FS} & \textbf{\# bio} \\
    \hline
    ChatGPT & 0.506 & 0.931 \\
    Llama-2-13b-chat & 0.736 & 0.908 \\
    Tulu-v2-13b-dpo & 0.755 & 0.827 \\
    \hline
    \hline
    Overall & 0.751 & 0.927 \\
    \hline
  \end{tabular}
  \caption{The per-sample Pearson correlation coefficient $r$ between the D-FActScore and the number of bios obtained by automatic evaluation and human evaluation.}
  \label{tab: correlation coefficient}
\end{table}

\subsection{Using ChatGPT to Extract and Group Atomic Facts}
\label{appendix: Using ChatGPT to Extract and Group Atomic Facts}
The automatic evaluation of D-FActScore relies on using GPT3.5 to extract and group atomic facts.
However, GPT3.5 has recently been deprecated (in mid Jan. 2024).
Here, we show that all our experiment results hold when replacing GPT3.5 with ChatGPT; i.e., we use ChatGPT to extract atomic facts from the paragraph and use ChatGPT to split atomic facts into groups.
We show the results in Table~\ref{table: subset result appendix}.
Comparing the result of using ChatGPT (Table~\ref{tab:human_eval_set_automatic_chatgpt appendix}) and using GPT3.5 (Table~\ref{tab:human_eval_set_automatic appendix}), we find that while the absolute numbers slightly changes, the observations stated in Section~\ref{subsection: Human VS Automatic Evaluation} still holds. 
We recap those results here:
\begin{itemize}
    \item Using ChatGPT or GPT3.5 in automatic evaluation shows the same factuality rankings among ChatGPT, Llama-13b-chat, and Tulu-v2-dpo as the ranking obtained with human annotation.
    \item D-FActScores obtained using ChatGPT and GPT3.5 are higher than human evaluation.
    \item Automatic evaluation can determine the number of biographies accurately.
    \item Open-source models lag behind ChatGPT.
    \item ChatGPT fully uses the received documents.
\end{itemize}

\begin{table}[t]
    \centering
    \begin{subtable}{\columnwidth}
        \centering
        \begin{tabular}{ccccc}
        \hline
         \textbf{Model}&  \textbf{FS}&  \textbf{D-FS}&  \textbf{\# bio}& \textbf{\# ent.}\\
         \hline
         ChatGPT&  98.3 &  92.1&  2.2 & 2.3 \\
         chat-13b&  94.8&  78.4&  1.0 & 1.7\\
         Tulu&  91.9&  83.2&  1.3 & 1.7\\
         
         \hline
        \end{tabular}
        \caption{Human evaluation}
    \label{tab:human_eval appendix}
    \end{subtable}
    \begin{subtable}{\columnwidth}
        \centering
        \begin{tabular}{ccccc}
        \hline
         \textbf{Model}&  \textbf{FS}&  \textbf{D-FS}&  \textbf{\# bio}& \textbf{\# ent.}\\
         \hline
         ChatGPT&  98.7&  96.3&  2.2& 2.3\\
         chat-13b&  95.3&  86.4&  1.1& 1.5\\
         Tulu&  95.8&  88.5&  1.3& 1.7\\
         
         \hline
        \end{tabular}
        \caption{Automatic evaluation (GPT3.5)}
    \label{tab:human_eval_set_automatic appendix}
    \end{subtable}
    \begin{subtable}{\columnwidth}
        \centering
        \begin{tabular}{ccccc}
        \hline
         \textbf{Model}&  \textbf{FS}&  \textbf{D-FS}&  \textbf{\# bio}& \textbf{\# ent.}\\
         \hline
         ChatGPT&  98.2 &  92.8 &  2.0 & 2.3\\
         chat-13b&  96.0&  87.1&  1.0& 1.5\\
         Tulu&  96.3 &  88.6&  1.2& 1.7\\
         
         \hline
        \end{tabular}
        \caption{Automatic evaluation (ChatGPT)}
    \label{tab:human_eval_set_automatic_chatgpt appendix}
    \end{subtable}
    \caption{FS: FActScore, D-FS: D-FActScore, \# bio: average number of separable biographies in one paragraph, \# ent.: average number of distinct entities in one paragraph, chat-13b: Llama-13b-chat.
    Human and automatic evaluations are conducted on the same set of paragraphs.
    The result in Table~\ref{tab:human_eval_set_automatic appendix} is obtained by using GPT3.5 to extract the atomic facts and split atomic facts into groups, and Table~\ref{tab:human_eval_set_automatic_chatgpt appendix} is the result of using ChatGPT to extract the atomic facts and split the atomic facts into groups.}
    \label{table: subset result appendix}
\end{table}

\section{Citation Recall}

\subsection{Definition}
\label{subsection: citation recall}
Citation recall assesses the citation quality when generating text with citations by measuring whether the generated text is fully supported by the cited documents.
For each sentence in the generation, its citation recall is $1$ if and only if the sentence has at least one citation and the sentence can be supported by its citation(s). 
\citet{gao-etal-2023-enabling} uses an NLI model to determine if the cited document supports the sentence.
The paragraph-level citation recall is the percentage of statements supported by its citations.
A high citation recall indicates that the generated content is well supported by the cited passages.

\paragraph{Shortcoming of Citation Recall}
While citation recall is specifically designed to evaluate the attribution of text with citations, the core concept of citation recall is very similar to FActScore, where a long-form generation is decomposed into claims, and each claim is verified independently.
As a result, citation recall also faces the same problem as FActScore in the case of entity ambiguity: even if citation recall says the paragraph is well-supported by the cited documents, the overall result can still be non-factual.  

\subsection{Experiment Result}
\label{subsection: appendix: citation recall}
We show the results of citation recall in Table~\ref{tab: citation recall}.
We discuss previouisly that citation recall cannot handle the non-factual paragraphs due to entity ambiguity.
As a result, even if the citation recall is high, the paragraphs can still be non-factual.
However, we do not see such a problem in the paragraphs generated in Section~\ref{section: Data preparation}.
This is because the citation recalls of the paragraphs generated by the LLMs are not very high, so there is no such a problem like "high citation recall but non-factual."
Nevertheless, we stress that citation recall cannot handle non-factual paragraphs due to entity ambiguity.

\begin{table}
\footnotesize
    \centering
    \begin{tabular}{cc}
         \hline
         & Citation Recall \\
         \hline
         ChatGPT & 56.65 \\
         Llama-chat-13b &  57.77\\
         Llama-chat-70b & 72.63 \\
         Tulu-v2-13b-dpo &  69.32 \\
         Vicuna-7b & 55.44\\
         \hline
    \end{tabular}
    \caption{The citation recall on 500 passages generated with 500 names in AmbigBio for each model.}
    \label{tab: citation recall}
\end{table}

\begin{table*}[h]
    \tiny
    \centering
    \begin{tabular}{|p{75em}|}
        \hline
        Please breakdown the following sentence into independent facts:\\
        \\
        Park Chan-wook, born on August 23, 1963, in Seoul, South Korea, is a renowned filmmaker and actor known for his impactful work in the film industry. He made his acting debut in the film "The Moon is the Sun's Dream" in 1992 and continued to appear in small and supporting roles throughout the 1990s.\\
\\
- Park Chan-wook was born on August 23, 1963.\\
- Park Chan-wook was born in Seoul, South Korea.\\
- Park Chan-wook is a renowned filmmaker.\\
- Park Chan-wook is an actor.\\
- Park Chan-wook is known for his impactful work in the film industry.\\
- He made his acting debut in the film.\\
- He made his acting debut in The Moon is the Sun's Dream.\\
- The Moon is the Sun's Dream is a film.\\
- The Moon is the Sun's Dream was released in 1992.\\
- After his acting debut, he appeared in small and supporting roles.\\
- After his acting debut, he appeared in small and supporting roles throughout the 1990s.\\
\\
Next, refer to the paragraph again and see if it explicitly states that it contains the biographies of multiple individuals. If there are multiple biographies, split the independent facts from different biography using \"- ===\". If the paragraph does not contain multiple biographies from different individuals, repeat the independent facts.\\
\\
- Park Chan-wook was born on August 23, 1963.\\
- Park Chan-wook was born in Seoul, South Korea.\\
- Park Chan-wook is a renowned filmmaker.\\
- Park Chan-wook is an actor.\\
- Park Chan-wook is known for his impactful work in the film industry.\\
- He made his acting debut in the film.\\
- He made his acting debut in The Moon is the Sun's Dream.\\
- The Moon is the Sun's Dream is a film.\\
- The Moon is the Sun's Dream was released in 1992.\\
- After his acting debut, he appeared in small and supporting roles.\\
- After his acting debut, he appeared in small and supporting roles throughout the 1990s.\\
\\
Please breakdown the following sentence into independent facts:\\
\\
Gavin Hamilton, born in Lanarkshire, Scotland in 1723, was a prominent neoclassical history painter and antiquarian who lived in Rome. He was also known for his role in the hunt for antiquities in the area. Gavin Hamilton, born in 1974, is an all-round cricketer who played for England in one Test and for Scotland in several One Day Internationals. Gavin Hamilton, who lived from 1561 to 1612, was the bishop of Galloway and was educated at the University of St. Andrews. Lastly, Gavin George Hamilton, born in 1872, was a Scottish Liberal politician and the 2nd Baron Hamilton of Dalzell.\\
\\
- Gavin Hamilton was born in Lanarkshire, Scotland in 1723.\\
- Gavin Hamilton was a prominent neoclassical history painter.\\
- Gavin Hamilton was a prominent antiquarian.\\
- Gavin Hamilton lived in Rome.\\
- He was known for his role.\\
- His role was in the hunt for antiquities.\\
- The hunt for antiquities was in the area.\\
- Gavin Hamilton was born in 1974.\\
- Gavin Hamilton is an all-round cricketer.\\
- Gavin Hamilton played for England in one Test.\\
- Gavin Hamilton played for Scotland in several One Day Internationals.\\
- Gavin George Hamilton was born in 1872.\\
- Gavin George Hamilton was a Scottish Liberal politician.\\
- Gavin George Hamilton was the 2nd Baron Hamilton of Dalzell.\\
\\
\\
Next, refer to the paragraph again and see if it explicitly states that it contains the biographies of multiple individuals. If there are multiple biographies, split the independent facts from different biography using \"- ===\". If the paragraph does not contain multiple biographies from different individuals, repeat the independent facts.\\
- Gavin Hamilton was born in Lanarkshire, Scotland in 1723.\\
- Gavin Hamilton was a prominent neoclassical history painter.\\
- Gavin Hamilton was a prominent antiquarian.\\
- Gavin Hamilton lived in Rome.\\
- He was known for his role.\\
- His role was in the hunt for antiquities.\\
- The hunt for antiquities was in the area.\\
- ===\\
- Gavin Hamilton was born in 1974.\\
- Gavin Hamilton is an all-round cricketer.\\
- Gavin Hamilton played for England in one Test.\\
- Gavin Hamilton played for Scotland in several One Day Internationals.\\
- ===\\
- Gavin George Hamilton was born in 1872.\\
- Gavin George Hamilton was a Scottish Liberal politician.\\
- Gavin George Hamilton was the 2nd Baron Hamilton of Dalzell.\\
        \hline
    \end{tabular}
    \caption{Two of the four demonstrations used for splitting the atomic facts from the paragraph based on the biography.}
    \label{tab:split bio prompt}
\end{table*}

\end{document}